\documentclass[conference, a4paper]{IEEEtran}

\ifCLASSINFOpdf
\else
\fi

\usepackage{graphicx}
\graphicspath{{./}}
\usepackage{amsmath}
\usepackage{url}
\usepackage[bookmarks=false]{hyperref}

\begin{document}

\title{Local Binary Pattern for Word Spotting in Handwritten Historical Document }

\author{\IEEEauthorblockN{
Sounak Dey\IEEEauthorrefmark{1},
Anguelos Nicolaou\IEEEauthorrefmark{1},
Josep Llados\IEEEauthorrefmark{1}, and
Umapada Pal\IEEEauthorrefmark{2}
}
\IEEEauthorblockA{
\IEEEauthorrefmark{1}
Computer Vision Center, Edifici O, Universitad Autonoma de Barcelona,Bellaterra, Spain
\\ 
\IEEEauthorrefmark{2}
CVPR Unit, Indian Statistical Institute, India  
\\
Email: \href{mailto:sdey@cvc.uab.es}{sdey@cvc.uab.es},
\href{mailto:anguelos@cvc.uab.es}{anguelos@cvc.uab.es}, 
\href{mailto:josep@cvc.uab.es}{josep@cvc.uab.es},
\href{mailto:umapada@isical.ac.in}{umapada@isical.ac.in}
}

}

\maketitle

\IEEEpeerreviewmaketitle

\begin{abstract}

Digital libraries store images which can be highly degraded and to index this kind of images we resort to word spotting as our information retrieval system. 
Information retrieval for handwritten document images is more challenging due to the difficulties in complex layout analysis, large variations of writing styles, and degradation or low quality of historical manuscripts.
This paper presents a simple innovative learning-free method for word spotting from large scale historical documents combining  Local Binary Pattern (LBP) and spatial sampling. 
This method offers three advantages: firstly, it operates in completely learning free paradigm which is very different from unsupervised learning methods, secondly, the computational time is significantly low because of the LBP features which are very fast to compute, and thirdly, the method can be used in scenarios where annotations are not available.
Finally we compare the results of our proposed retrieval method with the other methods in the literature.
\end{abstract}


\section{Introduction}

%
 
A lot of initiative has been taken to convert the paper scriptures to digitized media for preservation in digital libraries.
Digital libraries store different types of scanned images of documents such as historical manuscripts, documents, obituary, handwritten notes or records etc.
For several decades the information retrieval and computer vision community has been proposing techniques for indexing and retrieving huge amount of information in this imagery. 
The challenges in this area become diverse as more and more types of images are considered as input for archival and retrieval - for example historical letters or documents contain degraded information, bleed-through, historical handwriting etc. 

￼
%
Documents of different languages are also been archived in recent times which offers another challenge.
Traditional Optical Character Reader (OCR) techniques cannot be applied generally to all types of imagery due to several reasons.
In this context, it is advantageous to explore techniques for direct characterization and manipulation of image features in order to retrieve document images containing textual and other non-textual components.
A document image retrieval system asks whether an imaged document contains particular words, which are of interest to the user, ignoring other unrelated words. 
This is sometimes known as \textbf{keyword spotting} or simply 'word-spotting' with no need for correct and complete character recognition but by directly characterizing image document features at character, word or even document level. 
Word spotting technique in terms of pattern recognition can be defined as classification of word images.
 
 

The problem of word spotting, especially in the setting of large-scale datasets with millions to billions of images is balancing engineering trade-offs between number of documents indexed, queries per second, update rate, query latency, information kept about each  document and retrieval algorithm.
In order to handle such large scale data, computational efficiency and dimensionality is a critical aspect which is effectively taken care by the use of LBP in word spotting. 

To achieve information spotting in 
documents, several steps are necessary, which include noise removal, feature extraction, matching algorithm and indexing. 
We also come across some multilingual imaged documents which need language independent algorithms and alternative representations. 
Word spotting is fundamentally based on appearance based features. 
In this work we like to explore the textural  features as an alternative representation offering a richest description with minimal computational cost. 
Moreover the nature of the handwritten words suggests that there is a stable structural pattern due to the ascender and descenders in the words.
In this paper, our aim is to propose end-to-end method which can improve the performance for word spotting in handwritten historical document images. The specific objectives are presented as following:
\begin{enumerate}
\item Develop a word spotting method for large scale un-anotated handwritten historical data.
\item Apply texture feature like LBP to capture the fine grained information about the handwritten words which is computationally cheap. Converting the text to meta-information. 
\item Combine the spatial knowledge using a Quad tree spatial structure\cite{sidiropoulos2011content} for pooling.
\end{enumerate}

We use LBP as a generic low level texture classification features, that donot incorporate any assumptions specific to a task.
Out here we consider every text- block as a bi-modal oriented texture.
The rest of this paper is structured as follows.
In section \ref{soa}, we give an overview of the state of the art. 
Then we describe the architecture of the method in details in the section \ref{method}. 
We persent the different results to verify the findings in section \ref{experiments}. 
Finally in section \ref{conclusions}, we conclude with the contribution and ideas about future work regarding this framework.




\section{State of the art}
\label{soa}
\subsection{Taxonomy of Methods}
The state of the art word spotting techniques can be classified based on various criteria :
Depending on whether segmentation is needed i.e. segmentation-free or segmentation-based.
Based on possibility on learning : learning-free or learning-based, supervised/ unsupervised.
Based on usability : Query By example (QBE) or Query-By-String (QBS).
In the following section the popular techniques are surveyed in comparison to our proposed method.
\subsubsection{Segmentation-free or segmentation-based}
In the segmentation-based approach, there is a tremendous effort towards solving the word segmentation problem\cite{rath2003features}
\cite{bhardwaj2008script} \cite{ghosh2015sliding}.
 One of the main challenges of keyword spotting methods, either learning-free or learning-based, is that they usually need to segment the document images into words
  \cite{rath2003features}\cite{liang2012synthesised} or text lines \cite{frinken2012novel}
  using a layout analysis step. 
 In critical scenarios, dealing with handwritten text and highly degraded documents \cite{likforman2007text}\cite{louloudis2009text} segmentation is highly crucial.  
 Any segmentation errors have a cumulative effect on subsequent word representations and matching steps. 
 The work of Rusiñol et.al.\cite{rusinol2011browsing} avoids segmentation by representing regions with a fixed-length descriptor based on the well-known bag of visual words (BoW) framework \cite{csurka2004visual}. 
The recent works of Rodriguez et.al. \cite{rodriguez2012model} propose methods that relax the segmentation problem by requiring only segmentation at the text line level. 
In \cite{gatos2009segmentation}, Gatos and Pratikakis perform a fast and very coarse segmentation of the page to detect salient text regions.
The represented queries are in the form of a descriptor based on the density of the image patches.
Then, a sliding-window search is performed only over the salient regions of the documents using an expensive template-based matching.

\subsubsection{Learning-free or learning-based}
Learning-based methods, such as \cite{rodriguez2012model} \cite{fischer2012lexicon} \cite{frinken2012novel}, use supervised machine learning techniques to train models of the query words.
On the contrary, learning-free methods, with dedicated matching scheme based on image sample comparison without any necessary associated training process \cite{rath2003features} \cite{leydier2009towards}. 
Learning-based methods are preferred for applications where the keywords to spot are a priori known and fixed.
If the training set is large enough they are usually able to deal with multiple writers. However, the cost of having a useful amount of annotated data available might be unbearable in most scenarios. 
In that sense methods running with few or none training data are preferred. 
Learning-based methods \cite{perronnin2009fisher}
\cite{thomas2010information}  employ statistical learning methods to train a keyword model that is then used to score query images.
A very general approach was recently introduced in \cite{perronnin2009fisher}, where the learning-based approach is applied at word level based on Hidden Markov Models (HMMs).
The trained word models are expected to show a better generalization capability than template images. 
However, the word models still need a considerable amount of templates for training and the system is not able to spot out-of-vocabulary keywords.
In the above work holistic word features in conjunction with a probabilistic annotation model is also used.
In \cite{fischer2012lexicon} Fischer et. al. used nine features.
 The first three were the features regarding the cropped window (height, width and center of gravity) and the rest were the geometric features of the contours of the writing.
Peronin et.al.\cite{perronnin2009fisher} presented a very general learning-based approach at word level based on local gradient features.
In Rodriguez et. al. \cite{rodriguez2008local} uses SIFT feature descriptor in their pipeline for word spotting in historical documents. 
In our case we use texture descriptor like Local Binary Pattern to do word spotting which is much faster and can be calculated at the run-time. 
In this paper we use the LBP for the first time to do a fast learning free word spotting schematic.
The learning free method unlike unsupervised methods (such as learning on first few pages and applying the method on the remaining dataset) can be used without any kind of tuning to any database.

\subsubsection{Query By example (QBE) or Query-By-String (QBS) }
The query can be either an example image (QBE) or a string containing the word to be searched (QBS). 
In query-by-string (QBS) approaches, character models – typically Hidden Markov Models (HMMs) – have been pre-trained. 
At query time the models of the characters forming the string are concatenated into a word-model.
QBS and QBE approaches have their own advantages and disadvantages. 
QBE approaches require examples of the word to be spotted which is not the case of QBS. 
On the other hand, QBS approaches require large amounts of labeled data to train character models which is not the case of QBE. 
The work of Almazan \cite{almazan2012efficient}, and of  Rusinol et.al. \cite{rusinol2011browsing} where the word images are represented with HOG descriptors and SIFT descriptors aggregated respectively, can successfully be applied in a retrieval scenario.
 Most of the popular methods either work on QBE or in QBS and the success in one paradigm cannot be replicated in another, as comparison between images and texts is not well defined. In the following, we focus on the QBE scenario. 

The LBP explained in the later section uses the uniformity to reduce the dimensionality to speed up the process of matching the feature vectors in the learning free paradigm unlike other state of the art method.

\section{Proposed Method}
\label{method}
In this section the use of oriented gradient property of the LBP has been utilized
to develop a fast learning free method for information spotting for large scale document database where annonated data is unavailable.   
\subsection{End-to-end Pipeline Overview}
\begin{figure}[ht]
\centering
\includegraphics[width=.90\linewidth]{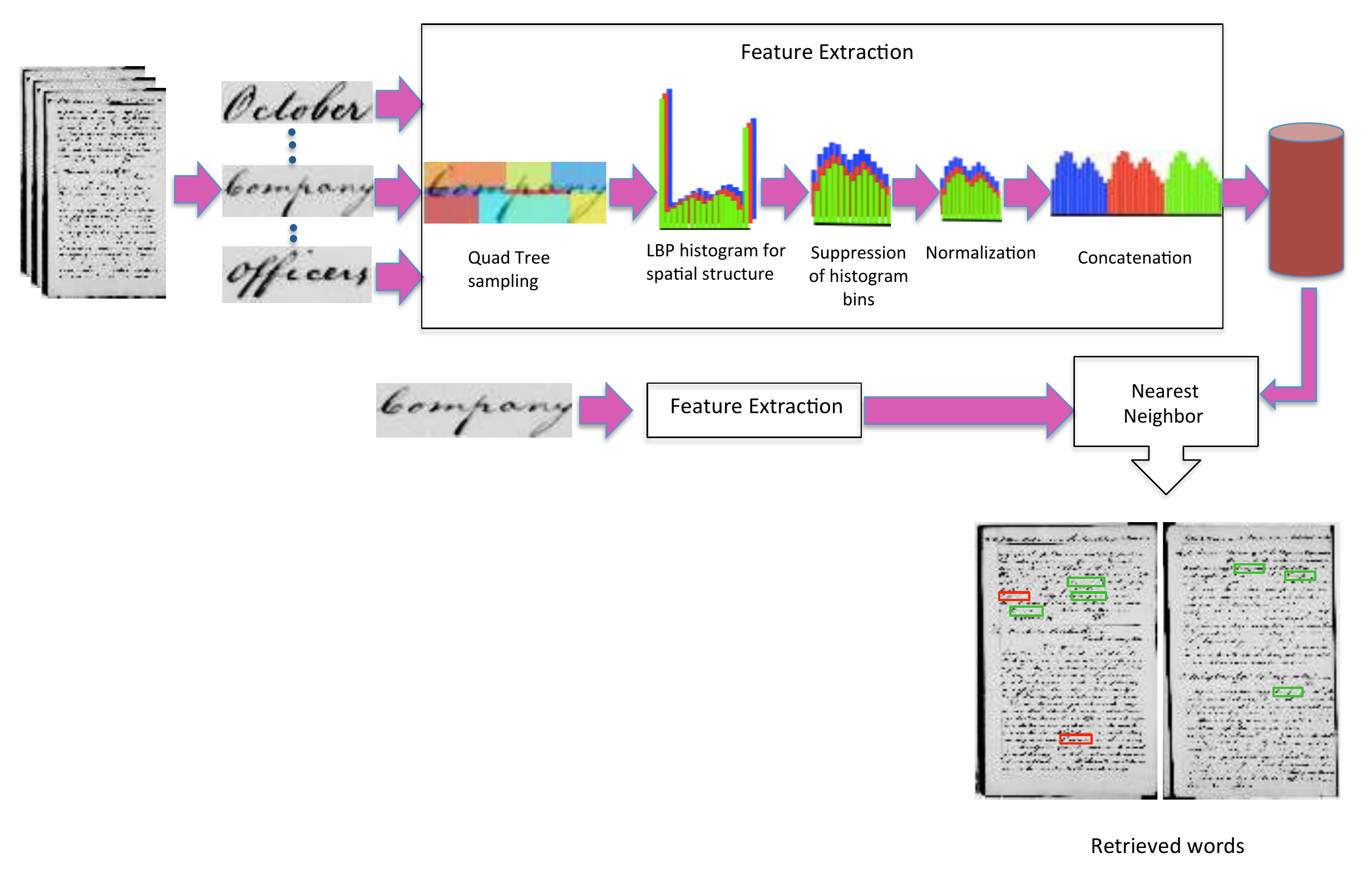} \\
\caption{Our proposed pipeline.}  
\label{fig: pipeline}
\end{figure}

In the pipeline given below we consider segmented words. The segmentation free approach can be facilitated by cascading the word segmentator as part as pre-processing technique in the pipeline. Our proposed pipeline is shown in Fig. \ref{fig: pipeline}. We use a median filtering preprocessing technique to reduce noise.\\

\begin{itemize}
\item \textbf{Quad Tree Spatial Sampling} : The gray level word images are then used to compute the spatial sub windows zones. 
This is done based on the center of mass of the image.
The center of mass of the whole image divides the image in four quadrants. 
Then each quadrant was further subdivided based on the center of mass of those quadrants.
This gave way to twenty such sub windows for the first two level.
The levels were experimentally fixed.  
The spatial information is embedded in the final feature vector using this technique.
LBP histogram is pooled over the zone created by this sampling technique as it gives more weightage to the zones having the pen strokes.
\\
\item \textbf{LBP Transform} : 
LBP is scale sensitive operator where the scaling depends on the sampling rate.
In our case the sampling rate is fixed as discussed below.
For each sub-window zone obtained in the spatial sampling state, a uniform compressed LBP histogram is generated.
The histogram of each such subwindow is then normalised and weighted by a edge pixel ratio in the sub-window.
This was perceived in this way, because the uniform LBP transform contains information regarding the sign transition of gradients which is prominent in case of the edges of the stroke width.
 Taking into account the number of edge pixel in the sub-window, in a way we gave importance to the sub-windows with more edge information.

The non-uniform pattern is suppressed to reduce the dimensionality of the final feature vector and also reduce its effect on normalization.
It can be seen in the Fig.\ref{fig: lbp}(d) the nonuniform patterns are purple in color while lying almost on the medial axis.
The information lost by suppressing the nonuniform pattern is very less compared to that of the uniform patterns.
The final feature is the concatenation of the histogram of each sub-window.
Though the dimensionality increases with respect to the number of level in spatial sampling the texture information becomes more distinctive for that space.
 \\
\item \textbf{Nearest Neighbor} : 
The feature thus obtained is compared to that of the query using Bray–Curtis dissimilarity matching as shown in eq. \ref{eq: braycurtis}.

\begin{equation}
\label{eq: braycurtis}
BC(a, b)=\frac{\sum\limits_{i}\mid a_i - b_i \mid}{\sum\limits_{i} a_i + b_i }
\end{equation}
where $a_i$ and $b_i$ are the $i$-th elements of the histograms.
We then use the width ratio which is the ratio between the width of the query and the images as an additional bit of information with the distance matrix.
The coefficient of the width ratio was experimentally decided.
Finally, the images with least distances are ranked chronologically. 
The performance of the system was measured by well established mean average precision, accuracy, precision and recall.
\end{itemize}

The vital steps of the pipeline are described in details in the following subsection.


\subsection{Local Binary Patterns}

The local binary pattern operator is an image operator, which transforms an image into an array or image of integer labels describing small-scale appearance of the image \cite{nicolaou2015sparse}.
  It has proven to be highly discriminative and its key points of interest, namely its invariance to monotonic gray level changes and computational profficiency, make it suitable for demanding image analysis tasks.
 The basic local binary pattern operator, introduced by Ojala et. al. \cite{ojala2002multiresolution}, was based on the assumption that texture has locally two complementary aspects, a pattern and its strength. 
LBP feature extraction consists of two principal steps: the LBP transform, and the pooling of LBP into histogram representation of an image.
As explained in \cite{ojala2002multiresolution} gray scale invariance is achieved because of the difference of the intensity of the neighboring pixel to that of the central pixel. 
 It also encapsulates the local geometry at each pixel by encoding binarized differences with pixels of its local neighborhood:

\begin{equation}
\label{eq: lbp}
LBP_{P,R,t}=\sum_{p=0}^{P-1} s_t(g_p-g_c)\times2^P,
\end{equation}

where $g_c$ is the central pixel being encoded, $g_p$ are $P$ symmetrically and uniformly sampled points on the periphery of the circular area of radius $R$ around $g_c$, and $s_t$ is a binarization function parameter by $t$. 
The sampling of $g_p$ is performed with bilinear interpolation.
$ t$, which in the standard definition is considered zero, is a parameter that determines when local differences are considered big enough for consideration.

In our LBP the original version of the local binary pattern operator works in a $3\times 3$ pixel block of an image. 
The pixels in this block are threshold by its center pixel value, multiplied by powers of two and then summed to obtain a label for the center pixel. 
As the neighborhood consists of $8$ pixels, a total of $2^8 = 256$ different labels can be obtained depending on the relative gray values of the center and the pixels in the neighborhood. 

\begin{figure}[ht]
\centering
\begin{tabular}{cc}
    \includegraphics[width=.4\linewidth]{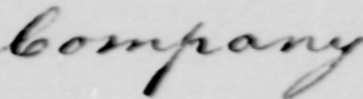} & \includegraphics[width=.4\linewidth]{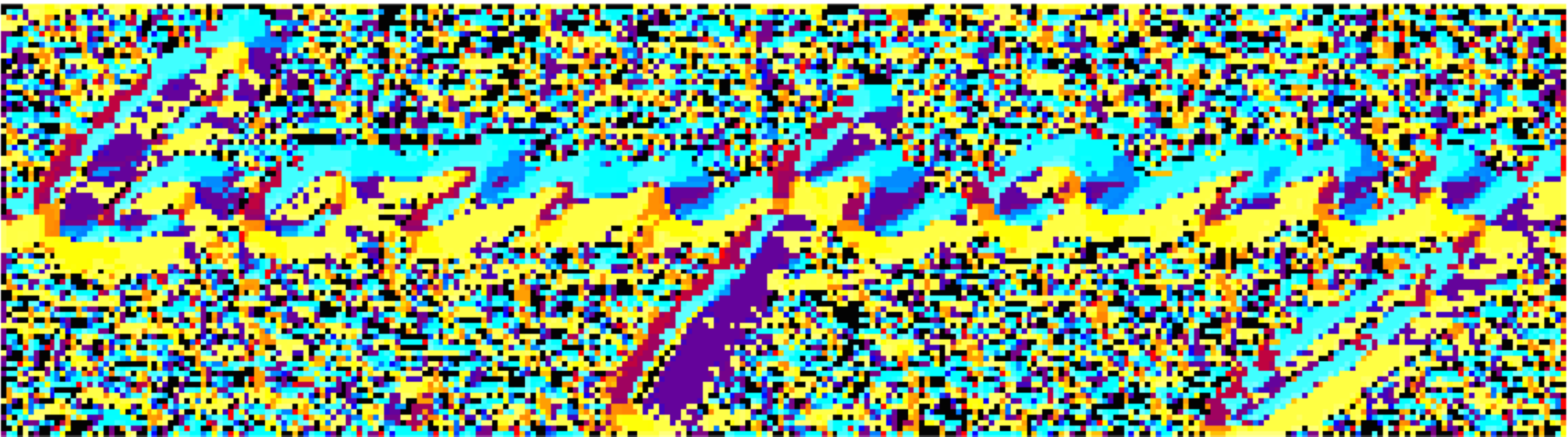}\\
    (a) & (b)\\
    \includegraphics[width=.4\linewidth]{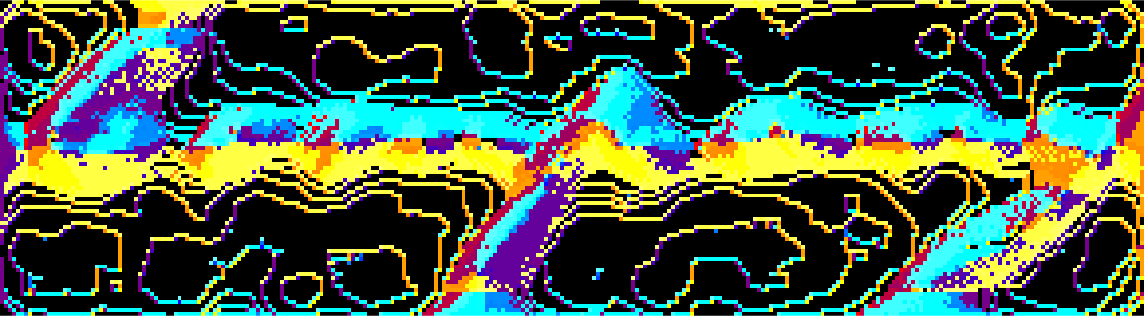} & \includegraphics[width=.4\linewidth]{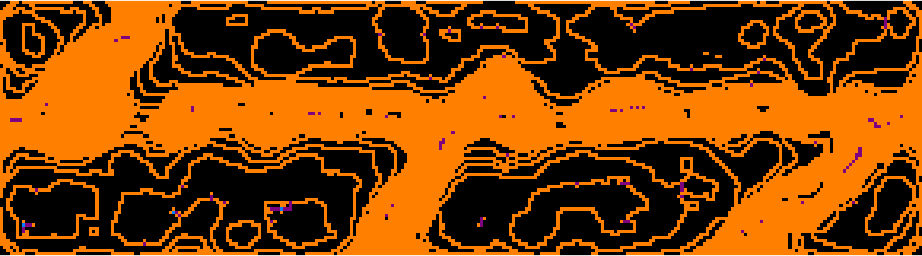}\\
    \\
  (c) & (d) \\
  \end{tabular}
\caption{(a) Input Image (b)LBP image (c)LBP with median filtering (d)LBP uniformity with median filter. } 
\label{fig: lbp} 
\end{figure}

In our case, we use the uniform LBP as mentioned in Ojala et.al. \cite{ojala2002multiresolution} which are the fundamental property of LBP, for the development of a generalized gray scale invariant operator Fig \ref{fig: lbp} (a). 
The term 'uniform' in case of local binary pattern refers to the uniformity of the appearance i.e. the circular presentation of the pattern has a limited number of transitions or discontinuities.
The patterns which are considered as uniform provide a vast majority over $90\%$, of the $3\times3$ texture patterns in the historical documents.
The most frequently observed 'uniform' patterns correspond to fundamental micro-features such as corners, spot and edges as shown in Fig \ref{fig: lbp} (b).
These are also considered as feature detectors for triggering the best pattern matching.
In our case, where $P=8$, $LBP_{8,R}$ can have $256$ different values.
We also designate patterns that have uniformity value of at most 2 as 'uniform' and uses the following Eq. \ref{eq: lbp_uniform}.
\begin{equation}
\label{eq: lbp_uniform}
LBP_{P,R}^{u2}=\left\{
  \begin{array}{ll}
  \sum\limits_{p=0}^{P-1} s(g_p-g_c) & \mbox{if}\   U(LBP_{P,R}) \leq 2  \\ \\
  P+1 & \mbox{otherwise} 
  \end{array}
  \right.
\end{equation}
where,
\begin{equation}
\begin{split}
\label{eq: lbp_pr}
U(LBP_{P,R})=\mid s(g_{P-1}-g_c)-s(g_0-g_c)\mid \\
+\sum_{p=1}^{P-1}\mid s(g_{p}-g_c)-s(g_{p-1}-g_c)\mid
\end{split}
\end{equation}

The $U$ value is the uniformity value for at most 2. 
 From definition exactly $P+1$ 'uniform' binary patterns can occur in a circularly symmetric neighbor set of $P$ pixels.
 Eq. \ref{eq: lbp_uniform} assigns a unique label to each of them, corresponding to the number of '1' bits in the pattern $(0\rightarrow P)$, while the 'nonuniform' patterns are grouped under the miscellaneous label (P+1).
 The final texture feature employed in texture analysis is the histogram of the operator outputs (i.e. pattern labels) accumulated over a texture sample.
  The reason why the histogram of 'uniform' patterns provides better discrimination in comparison to the histogram of all individual patterns comes down to differences in their statistical properties Fig. \ref{fig: lbp}(d).
   The relative proportion of 'nonuniform' patterns of all patterns accumulated into a histogram is so small that their probabilities can not be estimated reliably. 
\subsection{Spatial Sampling}

\begin{figure}[ht]
\centering
\begin{tabular}{cc}
    \includegraphics[width=.45\linewidth]{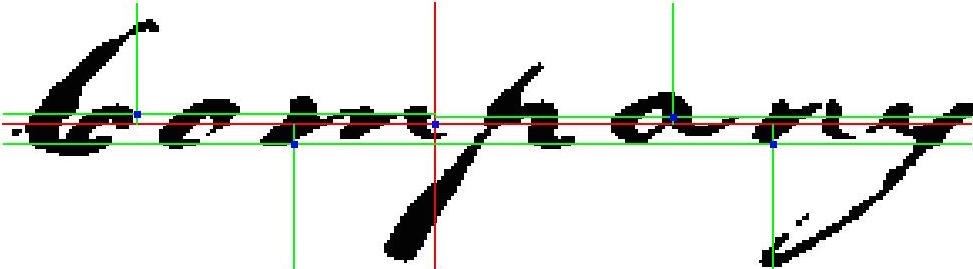} & 	
    \includegraphics[width=.45\linewidth]{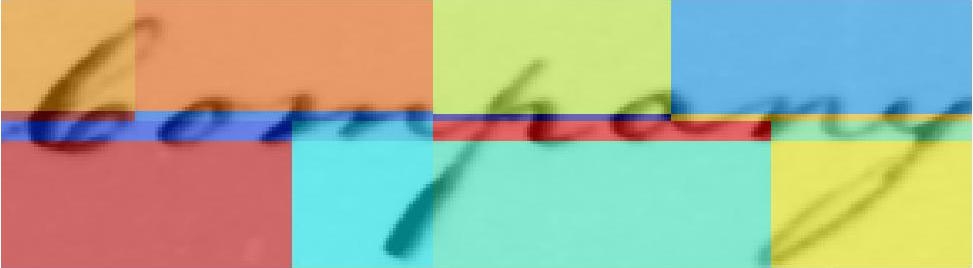}  \\
  (a) & (b) \\
  \end{tabular}
\caption{Spatial sampling using Quad tree technique. (a) Quad tree applied based on the center of mass where the red lines is the first level of sampling with blue point as center using suboptimal binarization. The green lines are second level of sampling with blue points as center respectively. (b) The different zones of the sampling is shown by different colors.} 
\label{fig: kdtree} 
\end{figure}

Since LBP histogram disregard all information about the spatial layout of the features, they have severely limited descriptive ability. 
We consider the spatial sampling as sub-windows on the whole image.
The sub-windows are obtain from the spatial pyramid with two levels.
This can be thought of as grid structure over the image.
The Quad tree image sampling based on the center of mass which yields much better results as shown in the Fig. \ref{fig: kdtree}.
The intuition was that the smaller the sub-window, with more black pixel density has more chances of having more discriminating power than the other sub windows.
The black pixel concentration suggests several handwritten letters together.  
To determine the sub windows, the gray images were considered for center of mass which was calculated on its binarized image obtained using Ostu's technique.
On this center point the image was divided into four quadrants.
The number hierarchical levels determines the total amount of sub-window that is used.
For level 1 it being 4 sub-windows and for level 2 it being 20 sub-windows i.e. 4+16 =20 (for each quadrant a further 4 quadrants was generated and so on).
In our case, we just used level 2 which was experimentally fixed.
\subsection{Nearest Neighbour k-NN}

Nearest neighbor search (NNS), also known as proximity search, similarity search or closest point search, is an optimization problem for finding closest (or most similar) points.
 Closeness is typically expressed in terms of a dissimilarity function: the less similar the objects, the larger the function values. 
 In our case it is the Bray-curtis Dissimilarity as defined in Eq \ref{eq: braycurtis}.
We made a very naive NN technique for our pipeline.

\section{Experiments}
\label{experiments}
\subsection{Experimental Framework}
Our approach is evaluated on two public datasets that we can find online: The George Washington (GW) dataset \cite{fischer2012lexicon} and the Barcelona Historical Handwritten Marriages Database (BHHMD)\cite{fernandez2014bh2m} as shown in Table \ref{tbl:dataset}. 
\begin{table}[h] 
\small
\centering
\caption{Comparison of the utilities of the different historic databases.}
\label{tbl:dataset}
\begin{tabular}{|r||c|c|c|c|}
\hline
\textbf{Method} & \textbf{Pages} & \textbf{Writers} & \textbf{Century} \\ \hline\hline
George Washington &20 & 1-2 & 18th \\ \hline
Marriage Licenses (Annotated)  & 40 & $>$2 & 15-16th \\ \hline
\end{tabular}
\end{table}

The proposed approach has been evaluated only in segmentation-based word spotting scenarios.
 We have used a set of pre-segmented words with the aim of comparing our approach with other methods in the literature with the aim of testing the descriptor in terms of speed, compacity and learning independence. 
 The results for all the methods considered all the words in the test pages as queries.

The used performance evaluation measures are mean average precision (mAP), precision and rPrecision.
Given a query, we label the set of relevant objects with regard to the query as \textit{rel} and the set of retrieved elements from the database as \textit{ret}. 
The precision and recall is defined in terms of \textit{ret} and \textit{rel} in Eq. \ref{eq: recallprecision}. 
rPrecision is the precision at rank \textit{rel}.
\begin{equation}
\label{eq: recallprecision}
Precision(P) =\frac{\mid ret \cap rel \mid}{\mid ret \mid},\   Recall=\frac{\mid ret \cap rel \mid}{\mid rel \mid}
\end{equation}
mAP is computed using each precision value after truncating at each relevant items in the ranked list.
For a given query, $r(n)$ is a binary function on the relevance of the $n$-th item returned in the ranked list.
The performance has been measured by the mean Average Precision (mAP), which is defined in Eq. \ref{eq: map}.
\begin{equation}
\label{eq: map}
mAP=\frac{\sum\limits_{n=1}^{\mid ret\mid}(P@n\times r(n))}{\mid rel\mid}
\end{equation}

\subsection{Results on George Washington Dataset}
The George Washington database was created from the George Washington Papers at the Library of Congress and has the following characteristics: 18th century, English language, two writers, longhand script, ink on paper.
The dataset was divided in 15 pages for train and validation and the last 5 pages for test.
Table \ref{tbl:resultsGW} shows the performance of our method compared to other methods in same paradigm. 
In the Table \ref{tbl:resultsGW} Quad Tree method is an adpataion of \cite{sidiropoulos2011content}. The best results in each category is higlighted.
 
\begin{figure*}[ht]
\centering
\begin{tabular}{c}
\includegraphics[width=.12\linewidth]{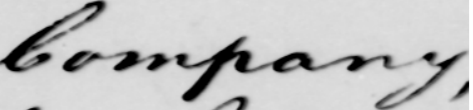}\\
\textbf{(a)}\\
    \includegraphics[width=.12\linewidth]{./company1.pdf} \includegraphics[width=.12\linewidth]{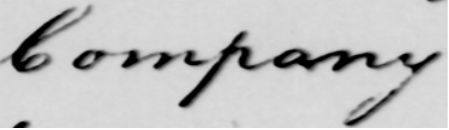}        		 \includegraphics[width=.12\linewidth]{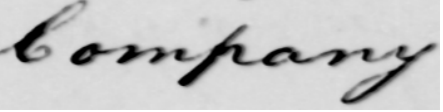} \includegraphics[width=.12\linewidth]{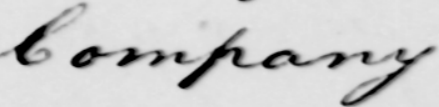} \includegraphics[width=.12\linewidth]{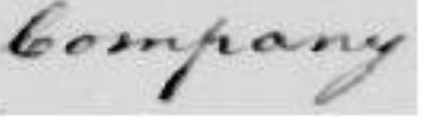} \\
 \includegraphics[width=.12\linewidth]{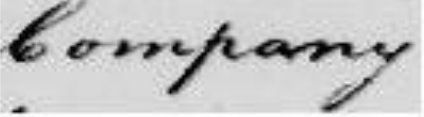} \includegraphics[width=.12\linewidth]{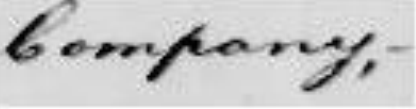} \includegraphics[width=.12\linewidth]{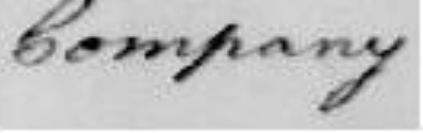} \includegraphics[width=.12\linewidth]{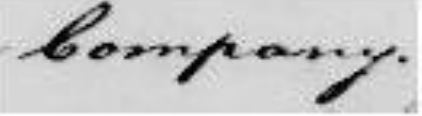} \includegraphics[width=.12\linewidth]{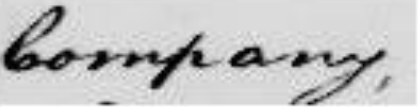} \\
 \includegraphics[width=.12\linewidth]{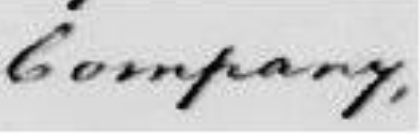} \includegraphics[width=.12\linewidth]{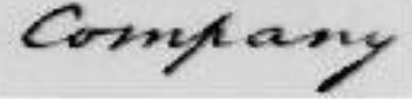} \includegraphics[width=.12\linewidth]{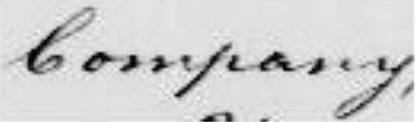} \includegraphics[width=.12\linewidth]{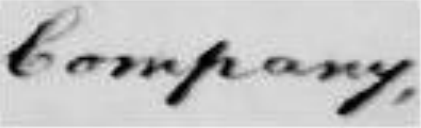} \includegraphics[width=.12\linewidth]{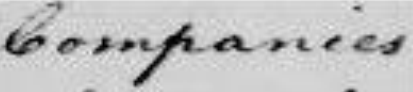} \\
 \includegraphics[width=.12\linewidth]{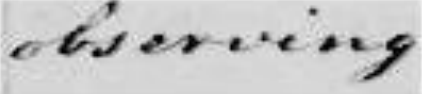} \includegraphics[width=.12\linewidth]{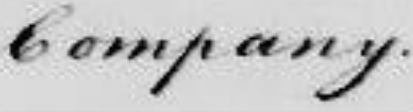} \includegraphics[width=.12\linewidth]{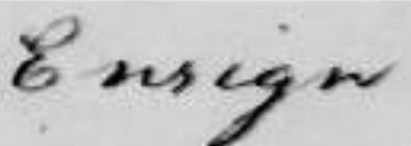} \includegraphics[width=.12\linewidth]{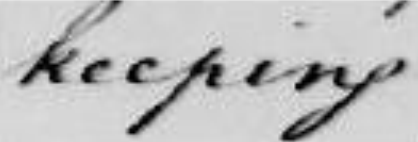} \includegraphics[width=.12\linewidth]{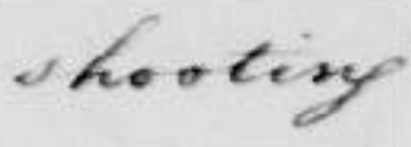} \\
 \textbf{(b)}\\
  \end{tabular}
\caption{Qualitative result examples. (a) Query (b) First 20 retrieved words.}  
\label{fig:results}
\end{figure*}

\begin{table*}[ht]
\small
\centering
\caption{Retrieval results on the George Washington dataset}
\label{tbl:resultsGW}
\begin{tabular}{|r||c|c|c|c|c|c|}
\hline
\textbf{Method}  & \textbf{Learning} & \textbf{mAP} & \textbf{Accuracy(P@1)} & \textbf{rPrecision} & \textbf{Speed(Test)} \\ \hline\hline

Quad-Tree & Standardization & $15.5\%$  & $30.14$  & $15.32$  & $44.41$secs  \\ \hline
BoVW \cite{rusinol2011browsing} & Unsupervised, codebook & $68.26\%$  & $85.87$  & $62.56$  & NA  \\ \hline
FisherCCA \cite{almazan2014segmentation} & Supervised & $\textbf{93.11\%}$  & $\textbf{95.44}$  & $\textbf{90.08}$  & $137.63$secs  \\ \hline\hline
DTW \cite{rath2003word} & No & $20.94\%$  & $41.34$  & $20.3$  & $78095.89$secs  \\ \hline
HOG pooled Quad-Tree & No & $48.22\%$  & $64.96$  & $43.04$  & $45.34$secs  \\ \hline
Proposed method (LBP) & No & $\textbf{54.44\%}$  & $\textbf{72.86}$  & $\textbf{48.87}$  & $\textbf{43.14}$secs \\ \hline\hline
\end{tabular}
\end{table*}
 Some qualitative results for an example query are shown in Fig. \ref{fig:results}.
  It is interesting to see that most of the words have been retrieved.
 This example takes the image \textit{Company} as a query. 
 The system correctly retrieves the first 15 words, whereas the the 16th retrieved word is \textit{observing}. 
 This also renders very similar to the query word in length and pattern. The next retrieved word is again a correct retrieval.
\subsection{Results on BHHMD Dataset}
This collection consists of registers of marriages in the Barcelona area between the 15th and 20th centuries.
The ground truthed subset contains 40 images.
The dataset was divided in 30 pages for train and validation and the last 10 pages for test.
Table \ref{tbl:resultsBarcelona} shows the performance of our method compared to other methods in same paradigm with respect to mean average precision.
Our method is the second best performing method with best computational time in this dataset.
We have also performed tests for cross dataset evaluation and our method in all category is just $2\%$ behind the best state of the art method. 
 In word spotting \cite{almazan2014segmentation} in mAP but performes exceptionally well in speed.

\begin{table*}[ht]
\small
\centering
\caption{Retrieval results on the BHHMD dataset}
\label{tbl:resultsBarcelona}
\begin{tabular}{|r||c|c|c|c|c|c|}
\hline
\textbf{Method}  & \textbf{Learning} & \textbf{mAP} & \textbf{Accuracy(P@1)} & \textbf{rPrecision} \\ \hline\hline

Quad-Tree & Standardization & $38.4\%$  & $61.92$  & $48.47$  \\ \hline
FisherCCA \cite{almazan2014segmentation} & Supervised & $\textbf{95.40\%}$  & $\textbf{95.49}$  & $\textbf{94.27}$  \\ \hline\hline
HOG pooled Quad-Tree & No & $66.66\%$  & $80.59$  & $62.35 $  \\ \hline
DTW \cite{rath2003word} & No & $7.36\%$  & $4.69$  & $2.99$  \\ \hline
Proposed method (LBP) & No & $\textbf{70.84\%}$  & $\textbf{84.13}$  & $\textbf{70.44}$  \\ \hline\hline
\end{tabular}
\end{table*}

\section{Conclusion}
\label{conclusions}
We have proposed a fast learning free word spotting method based on LBP-representations and a $k$-d tree sampling approach.
The most important contribution is that the proposed word spotting approach has been shown to be the best among the learning free ones in terms of performance.
The computational speed measured with the same benchmark for the proposed method is best compared to other state of the art methods.
We have shown that LBP based on uniformity can be stable under the deformations of handwriting.
For the pooling approach, the main contribution of the proposed framework is the pooling of the LBP based on the Quad Tree zones.
 The LBP has been defined as the textural feature which we utilize as oriented texture recognition.
  A sampling architecture has been designed to maximize the usage of the pen strokes and preserve the LBP patterns specific to the region.
  In terms of a retrieval problem, competitive mAP was obtained.
The time complexity of this indexation is linear in terms of the number of words in the database.
This was reduced to the order of $log N$ by using $k$-NN approach.
 It leads us to conclude that a feature extraction scheme as it is proposed here is very useful to compute inexact matchings in large-scale scenarios.
  We have demonstrated that compact textural descriptors are useful informations for handwriting word spotting, despite the variability of handwriting.
The experimental results demonstrates that our approach is comparable to other statistical approaches in terms of performance and time requirements.
 Future work will focus on the evaluation of the stability of LBP-based representations in large multi-writer document collections.

\section*{Acknowledgment}

This work has been partially supported by the Spanish project TIN2015-70924-C2-2-R, the European project ERC2010-AdG-20100407-269796 and RecerCaixa, a research program from ObraSocial ’La Caixa’.

\bibliographystyle{IEEEtran}

\bibliography{references.bib}

\end{document}